\documentclass[11pt,a4paper]{article}
\usepackage[hyperref]{acl2020}

\aclfinaltrue

\usepackage{times}
\usepackage{latexsym}
\usepackage{makecell}
\usepackage{subcaption}
\usepackage{verbatim}
\usepackage{listings}

\usepackage{footnote}
\makesavenoteenv{tabular}
\makesavenoteenv{table}

\usepackage{microtype}

\usepackage[ruled,vlined,linesnumbered]{algorithm2e}
\usepackage{amssymb}
\usepackage{graphicx}
\usepackage{amsmath}
\usepackage{booktabs}
\usepackage{bm}
\usepackage{mathrsfs}
\usepackage{ upgreek }
\usepackage{enumitem}
\usepackage{comment}
\usepackage{balance}

\newenvironment{packeditemize}{
\begin{itemize}
  \vspace{-1mm}
  \setlength{\leftmargin}{5pt}
  \setlength{\itemsep}{0pt}
  \setlength{\parskip}{0pt}
  \setlength{\parsep}{0pt}
  \vspace{-1mm}
}{\end{itemize}\vspace{-3mm}}

\usepackage{color}
\usepackage{soul}

\newcommand{\grace}[1]{{\bf \color{red} [[Grace says ``#1'']]}}

\title{High-Quality Diversification for Task-Oriented Dialogue Systems}

\author{Zhiwen Tang \qquad Hrishikesh Kulkarni \qquad Grace Hui Yang\\
InfoSense, Department of Computer Science\\
Georgetown University\\
\texttt{\{zt79,hpk8,grace.yang\}@georgetown.edu}
}

\begin{document}

\maketitle
    
\begin{abstract}

Many task-oriented dialogue systems use deep reinforcement learning (DRL) to learn policies that respond to the user appropriately and complete the tasks successfully. Training DRL agents with diverse dialogue trajectories prepare them well for rare user requests and unseen situations. One effective diversification method is to let the agent interact with a diverse set of learned user models. However, trajectories created by these artificial user models may contain generation errors, which can quickly propagate into the agent's policy. It is thus important to control the quality of the diversification and resist the noise. In this paper, we propose a novel dialogue diversification method for task-oriented dialogue systems trained in simulators. Our method, Intermittent Short Extension Ensemble (I-SEE),\footnote{I-SEE codes can be found at \url{https://github.com/smt-HS/I-SEE}.} constrains the intensity to interact with an ensemble of diverse user models and effectively controls the quality of the diversification. Evaluations on the Multiwoz dataset show that I-SEE successfully boosts the performance of several state-of-the-art DRL dialogue agents. 


\end{abstract}

\maketitle

\section{Introduction}


\begin{table*}[t]
    \centering \small
    \resizebox{0.99\textwidth}{!}{
    \begin{tabular}{p{5cm}|p{5cm}|p{5cm}}
        \toprule
       No Diversification &  Full Diversification & Quality-Controlled Diversifica. (I-SEE)\\
         \hline
User: Hi! I would like to book hotel Leverton House.\newline
System: Leverton House is a cheap hotel.\newline
User: What is the post of Leverton House?\newline
System: Post of Leverton House is cb58rs.\newline
User: What is the type of the hotel?\newline
System: It is a guesthouse.\newline
User: Where is it located?\newline
System: Leverton House is located at 732-734 new market road.\newline
User: What is the hotel phone number?\newline
System: You can call the hotel at 01223292094.\newline
User: In which area of the city is this hotel located?\newline
System: There are many attractions in the centre of the city. \colorbox{yellow}{[irrelevant]} \newline
User: Ok, bye!\newline
System: Goodbye!
        & 
User: Hi! I would like to book hotel Leverton House.\newline
System: Hotel Leverton House is located on the east side of the city.\newline
User: What is the post of the hotel?\newline
System: The post of hotel Leverton House is cb58rs.\newline
... \newline
System: ... Shall I go ahead and book it?\newline
User: Ok! I am not sure about the duration of stay or the number of people staying.\newline
System: Booking confirmed. Anything else? \colorbox{green!50}{[success]}\newline
User: I would also like to book a train from Peterborough to Cambridge arriving this Sunday at 15:15.\newline
System: Found a train arriving at 8:38. \colorbox{yellow}{[incorrect]}\newline
User: I want the one arriving at 15:15.\newline
System: Found a train arriving at 6:09. \colorbox{yellow}{[incorrect]}\newline
User: Ok, Bye!\newline
System: Have a good day!
        & 
User: Hi! I would like to book hotel Leverton House.\newline
System: Are you interested in booking hotel Leverton House on the east side?\newline
User: Where is it exactly located? What is the post, phone and type of the hotel?\newline
... \newline
System: ... Would you like to go ahead and book it?\newline
User: Ok, Not sure on the duration of stay or number of people as of now.\newline
System: Booking confirmed. \colorbox{green!50}{[success]}\newline
User: I would also like to book a train from Peterborough to Cambridge arriving at 15:15 on Sunday.\newline
System: Would you like to go ahead and book tickets on TR7721? \colorbox{blue!35}{[correct]}\newline
User: Ok. Not sure on the number of people.\newline
System: One ticket booked. \colorbox{green!50}{[success]}\newline
User: Great! Thanks!\newline
System: Have a good day!        \\
        \hline
        \textbf{Failure}: The system fails to answer the area question from the user and it does not book the hotel for the user. & \textbf{Failure}: Even though the system successfully answers all questions from the user and books the hotel, it fails to identify the correct train based on arrival time and does not book the train ticket.        & \textbf{Success}: Both hotel and train ticket have been successfully booked satisfying all constraints.  \\
        \bottomrule
    \end{tabular}
    }
    \caption{Example Dialogues.}
    \label{tab:Example2}
\end{table*}

Task-oriented dialogue agents assist human users to complete their tasks in multi-round human-agent interactions. Example tasks include booking a movie ticket or reserving a lunch table. Many agents use deep reinforcement learning (DRL) to learn good policies that respond appropriately in the dialogue and succeed in completing the task~~\cite{DBLP:journals/corr/SchulmanWDRK17,DBLP:conf/acl/TakanobuLH20,DBLP:conf/acl/GaoWPLL18}. Due to the high cost of interactive training using real human users, simulators have been used to replace the humans~\cite{DBLP:conf/emnlp/ShiQWY19}. 
Early user simulators rely on rules hand-crafted by domain experts, who review and summarize dialogue templates from everyday conversations~\cite{li2016user}. 
Rule-based simulators manage routine task scenarios efficiently; but they cannot easily generate unconventional and more diverse dialogue acts marked by human spontaneity.
Agents learning from these simulators may fail to handle rare queries and unseen situations. For example, in Table
\ref{tab:Example2}'s left column, the system learns from a rule-based simulator and fails to answer an unexpected question ``{\it In which area of the city is this hotel located?}" The user immediately ends the conversation and does not book the hotel.


\begin{figure}[t]
    \centering
    \includegraphics[width=\linewidth]{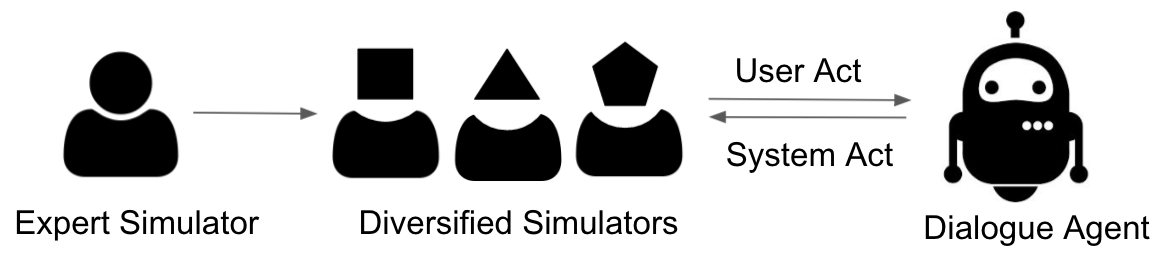}
    \caption{Agent Learns from Diversified Simulators.} 
    \label{fig:generalsystem}
\end{figure}

Increasing dialogue diversity is a long-lasting research interest motivated by different needs. In dialogue generation, diversity is desired to avoid dull responses. In simulator construction and reinforcement learning (RL), diversity is desired for training robust agents. Although the uses are slightly different, ideas to improve diversification can be universal. Dialogue diversity can be improved via i) enforcing diversity in objective functions (such as maximize mutual information) of neural models~\cite{li-etal-2016-diversity,baheti-etal-2018-generating}, ii) perturbing language rules~\cite{DBLP:conf/emnlp/NiuB19} or environment parameters~\cite{DBLP:conf/iros/TobinFRSZA17,DBLP:conf/iclr/RuizSC19}, iii) randomizing trajectory synthesis~\cite{DBLP:conf/nips/AndrychowiczCRS17,DBLP:conf/aaai/LuZC19}, iv) selecting more diverse data contributors~\cite{stasaski-etal-2020-diverse}, and v) sampling trajectories from a diverse set of  environments~\cite{DBLP:conf/nips/ChuaCML18,DBLP:conf/nips/JannerFZL19}. 

Figure \ref{fig:generalsystem} illustrates the last idea, which has received much attention recently~\cite{DBLP:conf/nips/ChuaCML18,DBLP:conf/nips/JannerFZL19}. This line of research diversifies  an agent's learning experiences by letting the agent interact with a diverse set of generative user models learned from an expert simulator. The idea involves little manual configuration and is often developed as model-based DRL (MBDRL)~\cite{DBLP:books/lib/SuttonB98}. 
MBDRL methods alternate between learning an environment  model and learning a policy. For a task-oriented dialogue agent, the environment model can be thought of a user model. It is a dynamic model updated to fit the trajectories the agent has collected so far; the policy then is optimized to maximize the expected long-term rewards within the model. Diversification of the user model is achieved by randomizing the parameter initialization of neural networks to our advantage. The agent, which is the policy learner, interacts with an ensemble of randomized user models to gain more diverse learning experiences.

However, one issue in this approach is that errors in (user) model learning may quickly propagate into policy learning. Table \ref{tab:Example2}'s middle column demonstrates a result from uncontrolled use of the diversified user models. In this example, even though the system successfully answers all questions from the user and books the hotel, the agent recommends two erroneous trains that do not satisfy the user's constraints and fail to book the ticket. This is because noise has been introduced to the training dialogues and they deviate too much from a legitimate conversation in real-life.


In this paper, we propose a novel dialogue diversification method, Intermittent Short Extension Ensemble (I-SEE), for task-oriented dialogues agents trained in simulators. First, I-SEE employs neural networks to learn a generative user model by imitating the expert simulator~\cite{DBLP:conf/ijcai/TorabiWS18}. Second, it  randomizes the parameter initialization of the neural networks to generate more user models, which  are diversification from the original expert-built simulator. These randomized user models form an ensemble of diverse simulators, named Diverse User Model Ensemble (DUME). Third, during policy learning, the agent interacts with multiple simulators to obtain diverse training trajectories. Particularly, we propose to mix trajectory segments sampled from the expert simulator and trajectory segments sampled from the DUME. This is to constrain the degree of noise introduced by diversification and do not divert too far from the expert simulator. Moreover, we propose to include the DUME trajectories only moderately frequently and for a short horizon. 

\begin{figure}[t]
    \centering
    \includegraphics[width=0.7\columnwidth]{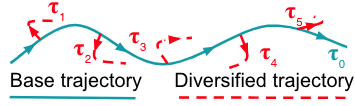}
    \caption{Conceptual illustration of I-SEE.}  
    \label{fig:conceptual}
\end{figure}

Figure \ref{fig:conceptual} illustrates our idea conceptually. By constraining the degree of diversification, I-SEE effectively controls the training trajectories' quality while preserving their diversity. In Table \ref{tab:Example2}'s last (right) example, 
the I-SEE agent successfully takes the booking task to a logical conclusion by correctly finding the TR7721 train, which satisfies the user's time constraints. We apply I-SEE to a few best performing DRL dialogue methods and evaluate them on the Multiwoz~\cite{budzianowski2018multiwoz} dataset. Results show that using DUME and I-SEE in combination would significantly improve the performance of these state-of-the-art systems. 

\section{Related Work} 

\subsection{Task-Oriented Dialogue Systems} 

Popular approaches for task-oriented dialogue systems include Sequence-to-Sequence (Seq2Seq) response generation~\cite{DBLP:journals/corr/VinyalsL15,DBLP:journals/corr/abs-2005-00796}, knowledge graph-driven question answering (KG-QA)~\cite{DBLP:journals/corr/abs-1910-03262,DBLP:conf/acl/MoonSKS19, DBLP:conf/aaai/YoungCCZBH18,DBLP:conf/acl/FungWM18,DBLP:conf/emnlp/MadottoCW0LLF20}, context-sensitive response retrieval~\cite{10.1145/3331184.3331265,DBLP:journals/corr/abs-2006-05009,Qu_2020,DBLP:journals/corr/abs-2101-06327} and RL~\cite{DBLP:journals/corr/BuckBCGHGW17,DBLP:conf/acl/GaoWPLL18,DBLP:journals/corr/abs-1912-00753,Luo:2014:WSD:2600428.2609629,DBLP:conf/ijcnlp/LiCLGC17,DBLP:conf/emnlp/SuLGLC18,DBLP:conf/acl/GaoWPLL18}. 

{\bf Seq2Seq dialogue agents} are generation methods. They use language models to capture the probability of one utterance given the previous, and based on the learned models to generate new utterances~\cite{DBLP:journals/corr/VinyalsL15,DBLP:journals/corr/abs-2005-00796}. These supervised methods take advantage of deep neural networks and infer effective encoder-and-decoders from large amount of sequential training data. Modeling the dialogue states~\cite{DBLP:conf/acl/CampagnaFML20} in the Seq2Seq architecture is a major interest in this line of research.  

{\bf KG-QA dialogue agents} enable reasoning and inference with pre-built knowledge graphs (KGs). The KGs can be about commonsense or domain-specific knowledge. A general KG can help a conversation more interesting and engaging \cite{DBLP:conf/acl/MoonSKS19, DBLP:conf/aaai/YoungCCZBH18}; while a specific KG can help accomplish the task more efficiently \cite{DBLP:conf/acl/FungWM18,DBLP:conf/emnlp/MadottoCW0LLF20}. Methods in this category focus on scaling up the KGs \cite{DBLP:conf/emnlp/MadottoCW0LLF20} and hopping mulitple steps on the KGs~\cite{DBLP:conf/acl/MoonSKS19}.    
 
{\bf Retrieval-based dialogue agents} leverage mature techniques in ad hoc retrieval and extend the techniques from individual queries to a session of them. Retrieval-based approaches do not rely on simulators; instead, learning from historical data, such as query logs, is still quite popular. This line of research focuses on revealing a user's mixed-initiative information need via asking back-and-force questions ~\cite{10.1145/3331184.3331265,DBLP:journals/corr/abs-2006-05009,Qu_2020,DBLP:journals/corr/abs-2101-06327}. 
However, when task complexity goes beyond the user's capability, these approaches may face difficulty in finding global solutions to the task goal. 


{\bf RL-based dialogue agents} can be grouped into model-free and model-based methods. Model-Free DRL (MFDRL) agents take a pre-built environment/simulator as it is and learn policies via direct interactions with it~\cite{DBLP:conf/ijcnlp/LiCLGC17,DBLP:conf/acl/DhingraLLGCAD17,DBLP:conf/ijcnlp/LiCLGC17,DBLP:conf/aaai/LiptonLG00D18,DBLP:conf/emnlp/SuLGLC18,DBLP:conf/emnlp/WuTG20}.  
On the contrary, model-based DRL (MBDRL) agents indirectly learn policies from the environment. MBDRL has two concurrent learning modules, namely model learning and policy learning. The {\it model learning} module can be thought of an additional computational layer between the environment and the agent. 
This provides opportunities to alter the original environment. 
MBDRL was originally proposed in robotics and control to speed up direct policy learning by inferring decision rules from past interactions and embedding them in the model. For dialogue agents, this middle layer of model learning acts as derived simulators (or learned user models) from the original expert simulator. Deep Dyna-Q (DDQ) \cite{DBLP:conf/acl/GaoWPLL18} is an MBDRL method built upon Dyna~\cite{DBLP:books/lib/SuttonB98}. D3Q \cite{DBLP:conf/emnlp/SuLGLC18} employs generative adversarial networks (GAN) to minimize the difference between trajectories generated from the learned models and that from the original expert simulator, assuming that the expert simulator is the gold standard. Likewise, ADC \cite{DBLP:conf/emnlp/WuTG20} uses double critics to mitigate the impact of poorly-generated trajectories to stabilize the agent's performance. Our method belongs to the family of MBDRL, with a focus on diversification. 


\subsection{Diversification in Dialogues}



Increasing dialogue diversity is a long-lasting research interest. Dialogue diversity can be improved via enforcing diversity objective functions (such as maximize mutual information) in neural models~\cite{li-etal-2016-diversity,baheti-etal-2018-generating},  perturbing language rules~\cite{DBLP:conf/emnlp/NiuB19} or environment parameters~\cite{DBLP:conf/iros/TobinFRSZA17,DBLP:conf/iclr/RuizSC19}, randomizing trajectory synthesis~\cite{DBLP:conf/nips/AndrychowiczCRS17,DBLP:conf/aaai/LuZC19}, selecting more diverse data contributors~\cite{stasaski-etal-2020-diverse}, and sampling trajectories from a diverse set of  environments~\cite{DBLP:conf/nips/ChuaCML18,DBLP:conf/nips/JannerFZL19}. For instance, \citeauthor{DBLP:conf/acl/CampagnaFML20} augmented dialogue data using domain-independent transition rules and domain-specific ontology~\cite{DBLP:conf/acl/CampagnaFML20}. \citeauthor{DBLP:conf/emnlp/NiuB19} synthesized more diverse dialogue trajectories by choosing semantic-preserving language perturbations via RL~\cite{DBLP:conf/emnlp/NiuB19}. 




\subsection{Diversification in DRL}


In model-free DRL, diversification can be achieved by domain randomization~\cite{DBLP:conf/iros/TobinFRSZA17,DBLP:conf/iclr/RuizSC19} or hindsight experience replay~\cite{DBLP:conf/nips/AndrychowiczCRS17,DBLP:conf/aaai/LuZC19}, without modeling the dynamics of the environment. 

In model-based DRL, diversification is done by altering the learned environment/user model; which are the closest to our work. For instance, \citeauthor{DBLP:conf/nips/ChuaCML18} proposed probabilistic ensemble trajectory sampling (PETS)~\cite{DBLP:conf/nips/ChuaCML18}, which learns an ensemble of environment models and uses them for planning. 
The follow-up work~\cite{DBLP:conf/nips/JannerFZL19} extended PETS with policy learning. 
Like us, \citeauthor{DBLP:conf/nips/JannerFZL19} concerned noise added by new trajectories generated by the derived environments. They proposed that the generation of new trajectories from the derived models should start from a beginning state shared with the original environment. These methods are mainly developed for robotics and work in continuous action space.   

In this paper, we propose to obtain mixed training trajectories by branching from the original trajectory generated by the expert simulator and extending with new trajectories by the derived simulators. Different from~\cite{DBLP:conf/nips/JannerFZL19}, our method is designed for dialogue agents' discrete action space. In our method, each training trajectory has an overlap much larger than \cite{DBLP:conf/nips/JannerFZL19} has with the expert trajectory. This allows us to obtain smoother transition distributions to facilitate discrete action space better. In addition, our method can parameterize the intensity to branch out, so that the level of diversification can be controlled and adjusted. 





\section{Problem Setup}

{\bf Task-Oriented Dialogue} is the interactive process between a user and a dialogue agent, who work together to accomplish a task. The process begins with the user initiating the dialogue with a task goal in mind. The task goal can have {\it constraints} and {\it requests}. Constraints are requirements a system response must satisfy and requests are for missing information the user needs to accomplish the task. E.g., a user wants to book tickets of a movie to be played on weekends but does not know the theater's phone number. Here the constraint is $time=weekend$ and request is $phone\_number=?$. The dialogue ends when both parties say ``good-bye" or the user abandons it.




{\bf Expert Simulator} 
is the rule-based simulator built by human experts.  It is denoted as $M_0$,  which describes how a typical user would choose proper dialogue acts as the dialogue unfolds.  The state of the expert simulator is $s^u_t$ at time step $t$ and the action is $a^u_t$ selected from an action set $A^u$, which can be either {\it making requests} or  {\it imposing constraints}. $M_0$ shows a mapping from $s^u_t$ to $a^u_t$, describing patterns and behaviours for the human users, and provides feedback to and converse with the dialogue agent. 


{\bf Diversified Simulator (or Diversified User Model) $M_{\phi_*}$} is a trainable user model that learns a parametric mapping from $s^u_t$ to ${a'}^u_t$ with parameter $\phi_*$. It mimics the behavior of the expert simulator $M_0$. With different parameter initialization, we can create a set of diversified user models. This set of diversified simulators is called Diversified User Model Ensemble ({\bf DUME}). 

{\bf Dialogue Agent (DA)} is the automatic response generator, who is expected to search in the knowledge base, reply the human users with relevant and correct answers, and make transactions following the user's requests. We use $s^s_t, a^s_t$ to denote the state and action of the dialogue agent at time step $t$. The agent also receives a reward signal $r_t$ as immediate feedback for its action $a^s_t$. Its state transition function $P$ models the probability of its next state given the current state and actions from both the user and the DA: $s^s_{t+1} = P(s^s_t, a^s_t, a^u_t)$. In the DRL setting, the DA is the policy learner. It learns a policy $\pi$ from a set of dialogue trajectories $\{\tau_*\}$. The goal of the agent is to learn a policy that can maximize the expected cumulative rewards $\mathbb{E}_{\pi} [\sum_t r_t]$ in a task-oriented dialogue. 
 
{\bf Interaction Tuple $\mathscr{T}$} is the state-action-reward tuple generated when the DA interacts with a simulator or a real user. At the $t^{th}$ dialogue turn, the $t^{th}$ interaction tuple is
 $\mathscr{T}_t = (s^s_t, a^s_t, r_t, s^u_t, a^u_t)$.

{\bf Trajectory Segment $\tau_j^k$} 
is a sequence of interaction tuples  when the DA interacts with a simulator ($M_0$ or $M_{\phi_i}$) or a real user, starting from time step $j$ to $k$: $\tau_j^k  = [\mathscr{T}_j, \mathscr{T}_{j+1+1}, ...,\mathscr{T}_k]$, 
where $\mathscr{T}_{t \in [j,k]}$ is the $t^{th}$ interaction tuple of the segment. Decided by the state transition function $P$, latter interaction tuples in $\tau$ depend on the earlier tuples. A {\bf base trajectory segment $\tau_0$} is a trajectory that records the interaction between the expert simulator $M_0$ and the DA. A {\bf diversified trajectory segment $\tau'$} is a trajectory segment that records the interaction between a diversified simulator $M_{\phi_*}$ and the DA. A {\bf full trajectory $\tau_0^{full}$=$[\mathscr{T}_0 ..., \mathscr{T}_{T}]$} starts from the beginning of a dialogue, s.t., $j=0$ and ends at $T$, where $T$ is the entire dialogue's length.  


\section{Proposed Work}

Our method aims to provide high-quality diversified training trajectories for task-oriented dialogue agents. 
We propose to (1)  construct an ensemble of diversified user models called DUME and (2) intermittently branching out short trajectories from the base trajectory using DUME and employ the new trajectories in policy learning. 

\begin{figure}[t]
    \centering
    \includegraphics[width=0.9\columnwidth]{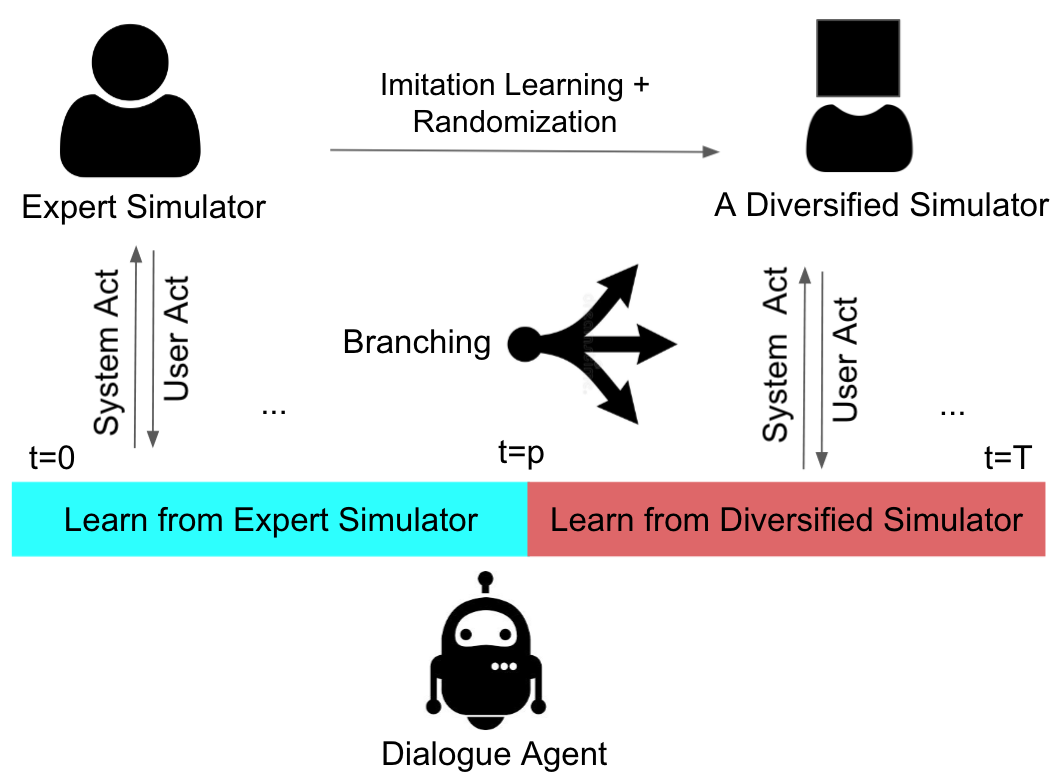}
    \caption{System Architecture} 
    \label{fig:isee_system}
\end{figure}

Figure \ref{fig:isee_system} illustrates the proposed system architecture. In our design, the dialogue agent can interact with both the expert simulator and a diversified simulator. Usually the agent starts with interacting with the expert simulator since $t=0$. At a branching step $t=p$, the agent switches to the diversified simulator to interact with, until the trajectory ends at $t=T$. The diversified simulator is obtained via imitation learning (from the expert simulator) and neural network initialization randomization. By controlling how frequently the branching should be performed and how long a diversified segment should be used, we effectively reach a balance between training data diversity and quality. 



\subsection{Constructing Diversified User Model Ensemble (DUME)}

To enhance dialogue diversity, we propose to have the agent interact with an ensemble of diverse user models $\{M_{\phi_*}\}$. We use neural networks with different initialization to learn diversified user models from the expert simulator $M_0$, and form the DUME using these learned models.

\subsubsection {Learning a single user model}
\label{sec:single}

We propose to learn the user models from the expert simulator by behavior cloning~\cite{DBLP:conf/ijcai/TorabiWS18}. For a single user model, we aim to learn a sequential decision-making function that maps $(s^u_1, s^u_2, ..., s^u_t...)$ to $ (a^u_1, a^u_2, ..., a^u_t, ...)$. The training inputs are from the base trajectories $\tau_0$, which includes a sequence of user state and user action pairs $\langle s^u_t, a^u_t\rangle$. The user state at the $t^{th}$ turn is 
\begin{equation}
    s^u_t = (G, \bigcup_{t'=1}^{t-1} a^s_{t'})
\end{equation}
where $G$ is the user goal, which can include both constraints and requests. $\bigcup_{t'=1}^{t-1} a^s_{t'}$ is the history of the dialogue agent's actions. 
The user action $a^u_t$ is 
\begin{equation} 
a^u_t=(a^u_{t, 1}, a^u_{t, 2}, ... , a^u_{t, |A^u|})
\end{equation}
where $a^u_{t, i}$ are binary variables indicating whether the $i^{th}$ dialogue act is active at dialogue turn $t$. $A^u$ are the available dialogue acts for the user. The ending of a dialogue is also a special dialogue act.  



Here a single user action can contain multiple dialogue acts. For instance, informing the destination and arrival time at the same dialogue turn when booking a train ticket. It means the number of dialogue acts per user action would vary. To allow the flexibility for modeling varied number of user acts, we propose to  break the training trajectory (which is a sequence) $\tau_0$ into individual state-action pairs and formulate the learning as choosing the right dialogue acts at a given state, i.e. learning the mapping from $s^u_t$ to $a^u_t$.  The optimization is done by minimizing the loss function $L(\phi)$:
\begin{equation}\label{eq:wm_loss}
\resizebox{0.8\linewidth}{!}{$
\begin{split}
    L(\phi) = - \frac{1}{|A^u|} \frac{1}{T} &  \sum_{t=1}^T \sum_{i=1}^{|A^u|}  a_{t, i} \log M_{\phi}(s_t^u)_i + \\
                                    & (1-a_{t,i}) \log (1 -  M_{\phi}(s_t^u)_i  )
\end{split}
$}
\end{equation}
where $\phi$ is the model parameter vector, $a_{t,i}$ is the ground truth indicator of whether the $i^{th}$ dialog act is taken at time step $t$, and $M_\phi(s_t^u)_i$ estimates the probability of the $i^{th}$ dialog act being chosen by the user model given $s^u_t$. The learning is performed by a multi-layer perceptron neural network parameterized by $\phi$. 




We are aware that the learning of the user models can be done using much more sophisticated methods. E.g., we can use more advanced neural network architectures and/or incorporate more information when defining the user states. However, these changes are not the main focus of this paper. The proposed user modeling is sufficient to support our investigation in exploiting them to improve diversification.  

\subsubsection{Forming a Diverse Ensemble} 



We propose to build an ensemble of diversified user models for better diversification. The ensemble, DUME,  contains a set of $E$ number of user models $M_{\phi_1}, M_{\phi_2}, ... , M_{\phi_E}$. Each of them is trained with behavior cloning as stated in Section \ref{sec:single}. DUME diversifies the user models by initializing the behavior cloning with different seeds. Each user model is trained using a separate neural network; these neural networks share the same architecture but use randomized, different initial parameters $\phi_j$. Our experiments (Section \ref{sec:exp_dume}) show that the diversity in DUME dramatically increases, as $E$ increases.





\subsection{Policy Learning with I-SEE}

\begin{algorithm}[t]
\SetAlgoLined
\small
\SetKwInOut{Input}{Input}
\SetKwInOut{Output}{Output}
\Input{Simulator $M$, \\
        Dialogue agent policy $\pi$, \\ Initial user state $s^u_0$,\\ Maximum trajectory length $T$}
\Output{
Dialogue trajectory dataset $D$\;
}
$D = \varnothing$\;
Initialize the user state to $s^u_0$\;
\For{$T$ time steps}{
        The user/simulator observes the state $s^u_t$ and takes action $a^u_t=M^i(s^u_t)$\;
        The agent observes the state $s^s_t$ and takes action $a^s_t = \pi(s^s_t)$\; 
        The agent receives reward $r_t$\;
        Store the interaction tuple $\langle s^s_t, a^s_t, r_t, s^u_t, a^u_t \rangle$ in $D$\;
        \If{the user/simulator decides to end the dialogue in $a^u_t$}{
            break\;
        }
    }
\Return $D$
\caption{Trajectory Generation} \label{algo:traj_gen}
\end{algorithm}

One would imagine that the more diversified trajectories used in training, the more robust the policy would be. An intuitive idea is to interact with the diversified user models $M_{\phi_*}$ from the  beginning to the end, without using $M_0$ at all. 
DDQ \cite{DBLP:conf/acl/GaoWPLL18} indeed exploits this design. However, a dialogue trajectory completely generated by $M_{\phi_*}$ suffers from accumulation of generation errors because they may  deviate too much from what a real conversation looks like. 

In this paper, we propose to learn from training trajectories generated from mixed sources. Our idea is to have controlled diversification during policy learning, where some of the learning is done by learning from the original expert simulator and some is done by learning from the diversified user models in DUME. The ratio of the diversified portion can be controlled as a hyper-parameter. The following details our method. 

\begin{algorithm}[t] 
\SetAlgoLined
\small
\SetKwInOut{Input}{Input}
\SetKwInOut{Output}{Output}
\Input{Simulator ensemble size $E$\\ Branching horizon $H$\\ Diversification ratio $\eta$}
\Output{Dialogue agent's policy $\pi$}
Initialize an ensemble of $E$ user models\;
Initialize the dialogue agent policy $\pi$\;
\While{ the dialogue agent's policy does not converge}{
 $D_{base}, D_{dvs} = \varnothing, \varnothing$ \;
  \For{every episode}{
  Initialize the expert simulator $M_0$\;
  Observe the initial user state $s^u_0$\;
  $D_{base}=$TrajectoryGeneration$(U, \pi, s^u_0, \infty)$\;
  }
  \While{$|D_{dvs}| < \eta |D_{base}|$}{
    Sample a simulator $M_{\phi_j}$ from the ensemble\;
    Sample a state $s^u_t$ from $D_{base}$ as the start state\;
    \begin{tabular}{l}
         $D_{dvs}= D_{dvs} \bigcup$ \\
         TrajectoryGeneration$(M_{\phi_j},  \pi, s^u_t, H)$\; \\
    \end{tabular}
  }
  Update the dialogue agent's policy $\pi$ with $D_{base} \bigcup D_{dvs}$ \; 
  Update the simulator ensemble with $D_{base}$ using Eq. \ref{eq:wm_loss}\;
}
\caption{Intermittent Short Extension Ensemble (I-SEE)} \label{algo:i-see}
\end{algorithm}

\subsubsection{Diversifying the Trajectories} 


During policy learning, the dialogue agent collects training trajectories generated from the simulators, to keep refining its policy based on gradient ascent. Algorithm \ref{algo:traj_gen} details the trajectory generation process. In order to sample a trajectory, the policy learner, i.e. the dialogue agent, interacts with a user model to obtain interaction tuples step by step and store each individual tuple in a dataset $D$. To obtain an individual interaction tuple, the simulator needs to take an action based on its own user model, and then the agent performs an action based on the state and its current policy $\pi$. The agent receives rewards and the next state from the simulator. The interaction tuple is stored and would be used later to form a full trajectory. This process works the same regardless the agent interacting with the expert simulator or  a diversified simulator. 

In this work, we propose to diversify the agent's learning experiences by learning from trajectories generated from mixed sources. First, we generate a full base trajectory $\tau_0^{full}$=$[\mathscr{T}_0 ..., \mathscr{T}_{T}]$ from the expert simulator and store all its tuples. Second, we pick a branching tuple $\mathscr{T}_p \in \tau_0^{full}$ at a branching point $p \in (0,T)$. Third, from $p$ onward, the trajectory is generated with a diversified user model $M_{\phi_*}$, which would take an action ${a'}^u_p$ different from the expert action $a^u_p$ and the agent would also land in a different state ${s'}^s_{p+1}=P(s^s_p, a^s_p, {a'}^u_p)$.

Such interaction with the diversified simulator $M_{\phi_*}$ continues with $H$ steps, resulting a diversified trajectory segment. The {\bf diversified trajectory segment $\tau_p'^{p+H}$} records the interaction between $M_{\phi_*}$ and the agent, extending the base trajectory $\tau_0$ from a branching point $p$ and running from $p+1$ onward. It is denoted as: 
\[
\tau_p'^{p+H} = [\mathscr{T}_p, \mathscr{T}_{p+1}', ... ,\mathscr{T}_{p+H}'], 
\]
where $p$ is the branching point and $p>0$, and $H$ is $\tau_p'$'s  horizon. The first interaction tuple in $\tau_p'$ is copied from the $p^{th}$ turn in $\tau_0$, i.e., $\mathscr{T}_p'= \mathscr{T}_{p}$.  The {\bf full trajectory with diversification} is thus $\tau_p^{full}=[\mathscr{T}_0, ..., \mathscr{T}_p, \mathscr{T}_{p+1}', ...,\mathscr{T}_{p+H}']$.

Our method generates parts of a dialogue with the diversified simulator and the other parts using the expert simulator. Each training trajectory  thus has overlaps with the expert trajectory, which obtains smoother transition distributions to facilitate the discrete action space that a dialogue agent has.





\subsubsection{Intermittent, Short Extensions} 

Further, we control the quality of diversification by using the DUME conservatively --  only use the DUME trajectories for a short horizon and intermittently -- to avoid accumulating generation errors. 

{\bf Branching Horizon.} The hyper-parameter $H$ is the branching horizon that controls how far a trajectory is generated from DUME. 
The larger the horizon $H$, the more diverse the resulting trajectory. Setting $H$ too small may cause the policy to be myopic as actions take time to show effects; whereas setting it too large may result in accumulation of errors. Our experiments show that using a moderately small $H=5$ is preferable. An analysis is reported in the experiment section.

{\bf Branching Intensity.} Another factor that determines the degree of diversification is the intensity of branchings. 
Instead of branching at every single step, our method only intermittently forks a diversified trajectory uniformly. 
This is done by setting a diversification ratio $\eta$ between the times the agent interacting with the expert simulator $M_0$ and with DUME. The diversification ratio $\eta$ is calculated as:  
\begin{equation}
    \eta = \frac{count(\mathscr{T}'_i, \forall i \in D_{dvs}) }{count(\mathscr{T}_j, \forall j \in D_{base})}
\end{equation}
where $\mathscr{T}'_i$ is a diversified interaction tuple stored in $D_{dvs}$ and $\mathscr{T}_j$ is an interaction tuple stored in $D_{base}$. $D_{base}$ and $D_{dvs}$ are collections of individual interaction tuples obtained as Lines 4-14 in Algo.~\ref{algo:i-see}. A larger $\eta$ means more diversified the agent's learning is.
Algo.~\ref{algo:i-see} shows the entire I-SEE algorithm.

\section{Experiments}

\subsection{Experimental Setup}

{\bf \textbullet \space Dataset.}
We evaluate the proposed approach on the Multiwoz \cite{budzianowski2018multiwoz} dataset. Multiwoz is a large-scale benchmark dataset for task-oriented dialogue systems. It has seven task domains, including restaurant, hotel, attraction, taxi, train, hospital and police. 
One dialogue may involve multiple task domains, which
is a good resemblance of how people converse in real life. Multiwoz 
provides 8,438 labelled dialogues, each dialogue of which is annotated by experts with a sequence of dialogue states and respective dialogue acts. Table \ref{tab:multiwoz_stat} shows the dataset statistics. 
The expert simulator in Multiwoz  starts a conversation and takes turns with a dialogue agent to dialogue.  The simulator may request information from the agent or give the agent permission to do new bookings. At each turn, the simulator or the agent can perform one or more dialogue acts. 
The agent is expected to 1) provide correct answers to requested information and 2) complete the booking, if asked.  

\begin{table}[t] 
    \centering 
   \resizebox{\linewidth}{!}{
    \begin{tabular}{rrrr}
        \#Domains & \#Dialogues &  Total \#Turns & \makecell{Avg \#Turns per dialogue} \\
         \hline
         7 & 8,438       &  113,556     &   13.46    \\
           \#Slots & \#Values & \makecell{Total DB Entries} & \makecell{Avg Entries per domain}\footnote{Five out of seven domains require querying the database.}  \\
        \hline
              24      & 4,510  & 3,116  & 623 \\
    \end{tabular}
   }
    \caption{Dataset Statistics (Multiwoz).} 
    \label{tab:multiwoz_stat}
\end{table}

\noindent {\bf \textbullet \space Evaluation Metrics.} 
{\bf Success} is our main metric, which is the success rate over all dialogue tasks tested. A task is successful if and only if 1) all the requested information is provided, and 2) all the booked entities match the user's requirements.  
{\bf Inform F1} evaluates whether an agent provides the information requested by the user. It is calculated as $F1=\frac{2 Prec * Recall}{Prec + Recall}$, where $Prec$  and $Recall$ are the precision and recall of the information replied by the agent. 
{\bf Match} evaluates whether the booked entities satisfy the user's requirement. It scores 1 if the correct entity is booked, otherwise 0. In the case of multiple bookings, the scores are averaged across all bookings. {\bf \#Turns} measures the number of turns a dialogue last regardless of its success. The less the turns, the better.
  
\noindent {\bf \textbullet \space Baselines.}
We compare the performance of a few top-performing DRL dialogue agents on the Multiwoz dataset with three settings. The settings are 1) the algorithm without diversification, 2) with full and uncontrolled diversification, and 3) with I-SEE. These baseline systems include state-of-the-art MFDRL and MBDRL methods and best performing DRL agents on Multiwoz. 
\textbf{DQN} (Deep Q-Network) ~\cite{DBLP:journals/nature/MnihKSRVBGRFOPB15} is an off-policy MFDRL method, which approximates the value function of state-action pairs with a deep neural network and learns the function using experience replay.
\textbf{PPO} (Proximal Policy Optimization)~ \cite{DBLP:journals/corr/SchulmanWDRK17} is an on-policy MFDRL algorithm, which optimizes a surrogate objective function which restricts the change of action distributions in a policy update. \textbf{GDPL} (Guided Dialogue Policy Learning)~ \cite{DBLP:conf/emnlp/TakanobuZH19} is the best performer on Multiwoz. It uses inverse RL to reconstruct reward function and optimizes its policy with PPO. 
\textbf{DDQ} (Deep Dyna-Q)~ \cite{DBLP:conf/acl/GaoWPLL18} is an MBDRL algorithm designed for task-oriented dialogue agents. DDQ generates complete trajectories from its environmental models, which is equivalent to our setting of DQN+full diversification.
\textbf{MADPL} (Multi-Agent Dialogue Policy Learning)~ \cite{DBLP:conf/acl/TakanobuLH20} is a multi-agent MFDRL method that trains the system and the user simulator simultaneously. It is also a leading performer on Multiwoz.  

\noindent {\bf \textbullet \space Implementation Details}
We use Multiwoz's agenda-based simulator  \cite{DBLP:conf/acl/ZhuZFLTLPGZH20} as the expert simulator. The DUME and policy networks and value networks in the baselines are learned using three-layer multi-layer perceptrons (MLPs). A learned user model has an input dimension of 230 and output of 67, with a hidden layer of 200 units. The DRL dialogue agents all use an the input layer of 553 units.  PPO's policy network uses a hidden layer of 200 units and output of 166. PPO's value network has a hidden layer of 50 and output of 1. DQN also uses a hidden layer of 200 units and output of 166. The I-SEE dialogue agent is trained with a mix of expert simulator and diversified simulators as presented in the paper and tested with only the expert simulator. 

\subsection{Effectiveness}

\begin{table*}[t]
    \centering
    \resizebox{\linewidth}{!}{
    \begin{tabular}{l|llllllll}
    \toprule
        {\bf Algorithm} &  {\bf Success} $\uparrow$ & {\bf Impr.} & {\bf Inform F1} $\uparrow$ & {\bf Impr.} & {\bf Match} $\uparrow$ & {\bf Impr.} & {\bf \#Turns} $\downarrow$ & {\bf Impr.\% } \\
    \midrule
         MADPL & 70.1 & &  76.26 & & 90.98 & &  8.96 & \\
    \midrule
        PPO     & 77.9 & &  86.45 & & 78.90 & & 9.785 & \\
        PPO+Dvs. & 69.0 & {\small (-8.9)} & 80.27 & {\small (--6.18)} & 70.55 & {\small (-8.35)} & 11.39 & {\small (-16.40\%)}  \\
        PPO+I-SEE & {\bf 84.5} & {\small \bf (+6.6, +15.5)} & {\bf 88.91} & {\small  \bf (+2.46, +8.64)} & {\bf 86.29} & {\small  \bf (+7.93, +15.74)}  &{\bf 8.88} & {\small  \bf (+9.25\%, +22.04\%)} \\
    \hline
        DQN     & 74.4 & & 87.61 & & {\bf 92.91} \footnotesize{best} & & 11.54 &  \\
        DQN+Dvs. (DDQ)  & 72.1 & {\small (-2.3)}  & 84.26 & {\small (-3.35)} & 82.04 & {\small (-10.87)} & 11.78 & {\small (-2.08\%)}\\
        DQN+I-SEE & {\bf 85.2} & {\small  \bf (+10.8, +13.1)}  & {\bf 90.18} & {\small  \bf (+2.57, +5.92)} & 92.59 & \small{\bf (}-0.32{\bf , +10.55)} & {\bf 9.83} & {\small \bf (+14.82\%, +16.55\%)}\\
    \hline
      GDPL  & 86.5 &  & \textbf{94.97} \footnotesize{best} & & 83.90 & & 7.64 & \\
      GDPL+Dvs. & 72.8 & {\small (-13.7)} & 80.86 & {\small (-14.11)} & 81.10 & {\small (-2.80)}  & 9.98 & {\small (-30.63)} \\
      GDPL+I-SEE & \textbf{93.2} \footnotesize{best}& {\small  \bf (+6.7, +20.4)} & 91.83 & \small{\bf (}-3.14{\bf , +10.97)} & \textbf{92.76} & {\small  \bf (+8.86, +11.66)} & \textbf{7.32} \footnotesize{best}& {\small  \bf (+4.19\%, +26.65\%)} \\
    \bottomrule
    \end{tabular}
    }
    \caption{Dialogue Effectiveness on Multiwoz. X+Dvs shows the improvement w.r.t. a baseline X. X+I-SEE reports the improvements w.r.t. X and X+Dvs, respectively.}  
    \label{tab:exp_main}  
\end{table*}

 

\begin{figure*}
    \centering
    \begin{subfigure}{0.24\textwidth}
    \centering
    \includegraphics[width=\textwidth]{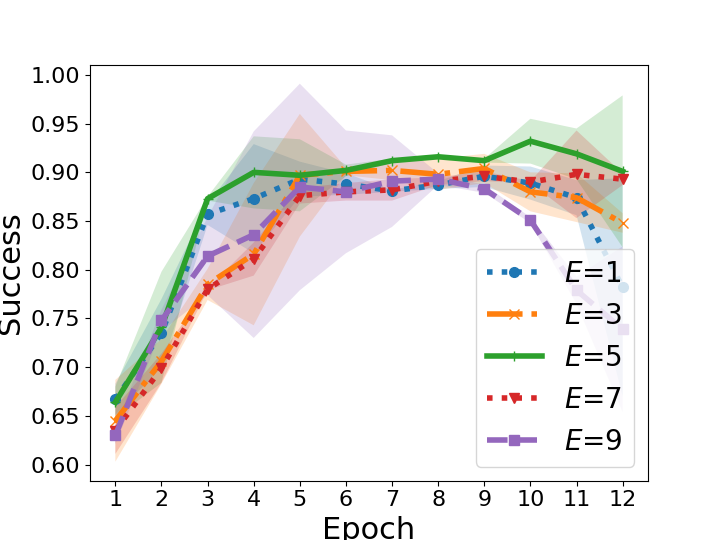}
    \caption{Learning curves of different ensemble size ($E$)}
    \label{fig:exp_ensemble}
    \end{subfigure}
    \begin{subfigure}{0.24\textwidth}
    \centering
    \includegraphics[width=\textwidth]{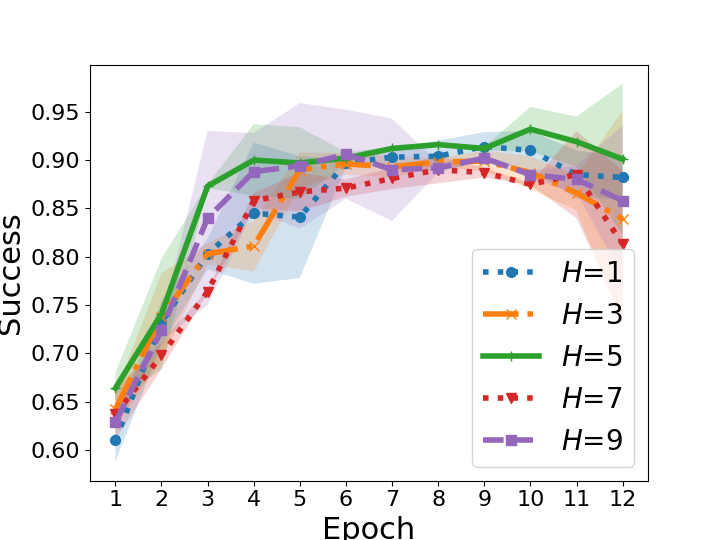}
    \caption{Learning curves of different diversification horizons ($H$)}
    \label{fig:exp_horizon}
    \end{subfigure}
    \begin{subfigure}{0.24\textwidth}
    \centering
    \includegraphics[width=\textwidth]{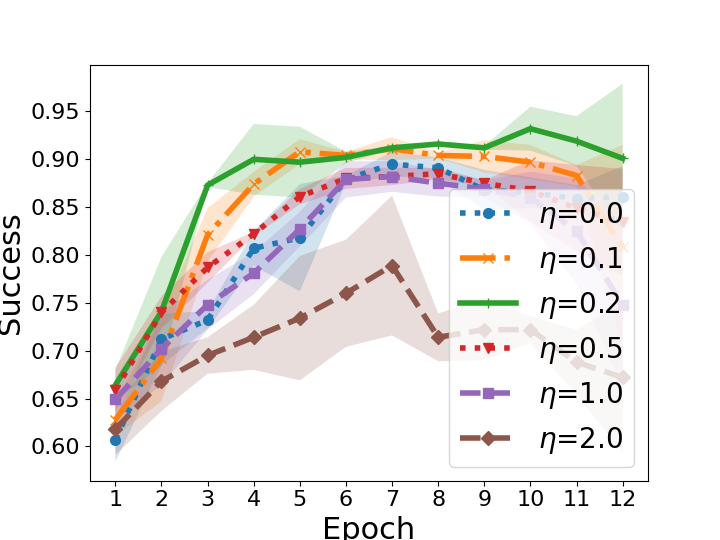}
    \caption{Learning curves of different diversification ratios ($\eta$)}
    \label{fig:exp_ratio} 
    \end{subfigure}
    \begin{subfigure}{0.24\textwidth}
    \centering
    \includegraphics[width=\textwidth]{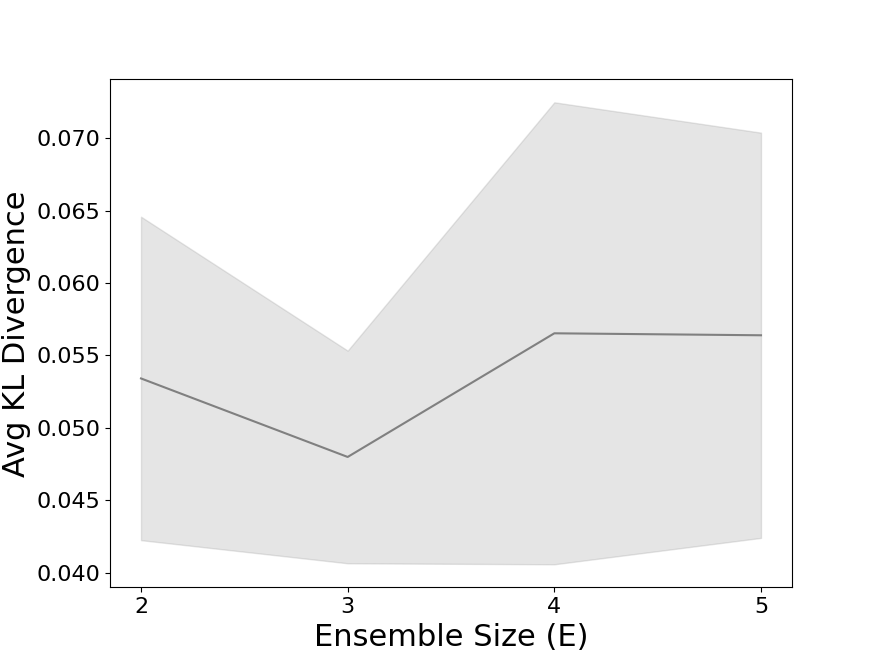}
    \caption{Evaluating the Diversity of DUME}
    \label{fig:diversity}
    \end{subfigure}
    \caption{Experiment Results on Multiwoz.} 
    \label{fig:my_label}
\end{figure*}

Table \ref{tab:exp_main} presents the experiment results. The proposed method
I-SEE outperforms the original algorithms and the full diversification variants for all baselines on the main metric, success, and the number of turns. The best performance is given by GDPL+I-SEE, with a success rate of 93.2 and only 7.32 dialogue turns on average. Moreover, the I-SEE variants  perform the best on {\it Inform F1} for 
PPO and DQN, and on {\it Match} for PPO and GDPL. The improvements are large. 
These results suggest that diversification in general improves a DRL dialogue agent's effectiveness. However, full and uncontrolled diversification may worsen the performance; while a moderate level of diversification as we propose is a better choice. 

\subsection{Analysis of I-SEE}

To understand why I-SEE works, we investigate the relationship between the degree of diversification and the success rate.  GDPL is selected as the baseline system $X$. We study three I-SEE hyper-parameters that are responsible for the degree of diversification. They are the user model ensemble size $E$, branching horizon $H$, and diversification ratio $\eta$. As each of these parameters gets bigger, the degree of diversification increases. We plot the dialogue agent's learning curves w.r.t the three parameters in Figures \ref{fig:exp_ensemble}, \ref{fig:exp_horizon}, and \ref{fig:exp_ratio}, respectively.

We observe that a single optimum exists for each hyper-parameter when they reach the best success rate. 
As we increase the size of the ensemble with $E=1,3,5,7,9$, the degree of  diversity increases. Figure \ref{fig:exp_ensemble} shows that initially increasing the diversity helps improve the performance; However, the trend turns downwards after reaching the optimum when $E=5$. Figures \ref{fig:exp_horizon} and \ref{fig:exp_ratio} demonstrate similar trends. In the end, the best combined I-SEE setting is $E=5, H=5,$ and $\eta=0.2$. This experiment suggest that diversification can only help an agent's learning to a certain extent; Too much diversification beyond that may introduce too much noise in the learning and hurt the agent's performance. Therefore, the degree of diversification must be carefully chosen in practice. 

\subsection{Analysis of DUME}
\label{sec:exp_dume}

DUME is our collection of trainable diversified user models. 
We calculate the average pairwise KL-divergence for every two models $M_{\phi_i}$ and $M_{\phi_j} \in $ DUME to directly measure the degree of diversity within DUME. Each user model  
 is run on the same stavte sequence $\{s^u_1, ..., s^u_t ...\}$ and outputs an action sequence $\{{a'}^u_1, ..., {a'}^u_t ...\}$. Since each ${a'}^u_t$ may contain multiple dialogue acts, we break down every ${a'}^u_t$ into individual dialogue acts and calculate the distribution over the dialogue act set $A^u$. The mean $\mu$ and standard deviation $\sigma$ of the KL divergences are plotted in Figure \ref{fig:diversity}. We can see that as DUME has bigger size, both $\mu$ and $\sigma$ increase; which means the differences between the DUME simulators dramatically increase and they would add much diversity into the agent's learning. 


\section{Conclusion}



This paper presents Intermittent Short Extension Ensemble (I-SEE), a DRL diversification method that successfully improves dialogue diversity and policy robustness while maintaining high data quality.
I-SEE uses an ensemble of trainable user models to achieve diversity and  controls the diversification quality by branching from original dialogue trajectories only for a short horizon and intermittently. 
Our experiments on Multiwoz show that using I-SEE  can significantly improve several best state-of-the-art DRL dialogue agents.


\section*{Acknowledgment}

This research was supported by the U.S. National Science Foundation Grant no. IIS-1453721. Any opinions, findings, conclusions, or recommendations expressed in this paper are of the authors, and do not necessarily reflect those of the sponsor.

\bibliographystyle{acl_natbib}
\bibliography{reference}

\balance

\end{document}